\documentclass[letterpaper, 10 pt, conference]{ieeeconf}  

\IEEEoverridecommandlockouts                              

\usepackage{graphicx}

\usepackage{amsmath,amsfonts}
\usepackage{algorithmic}
\usepackage{algorithm}
\usepackage{array}
\usepackage{textcomp}
\usepackage{stfloats}
\usepackage{url}
\usepackage{verbatim}
\usepackage{graphicx}
\usepackage{multirow}
\usepackage{subcaption}
 \usepackage{booktabs} 
\usepackage{makecell}
 
\usepackage{lipsum}
\let\labelindent\relax
\usepackage{enumitem}
\usepackage{hyperref}

\title{\LARGE \bf
Learning Robust, Agile, Natural Legged Locomotion Skills in the Wild
}

\author{Yikai Wang$^{1}$\textsuperscript{\textdagger}, Zheyuan Jiang$^{2}$\textsuperscript{\textdagger} and Jianyu Chen$^{3}$
\thanks{\textsuperscript{\textdagger}Joint first authors}
\thanks{$^{1}$Yikai Wang is with Weiyang College, Tsinghua University, Beijing, 100084, China
        {\tt\small wangyiji20@mails.tsinghua.edu.cn}}%
\thanks{$^{2}$Zheyuan Jiang is with the Institute for Interdisciplinary Information Sciences, Tsinghua University, Beijing, 100084, China
        {\tt\small zy-jiang21@mails.tsinghua.edu.cn}}%
\thanks{$^{3}$Jianyu Chen is with Institute for Interdisciplinary Information Sciences, Tsinghua University, Beijing, 100084, China, and is also with the Shanghai Qi Zhi
Institute, Shanghai 200232, China{\tt\small jianyuchen@tsinghua.edu.cn}}%
}

\begin{document}

\maketitle
\thispagestyle{empty}
\pagestyle{empty}

\begin{abstract}

Recently, reinforcement learning has become a promising and polular solution for robot legged locomotion. Compared to model-based control, reinforcement learning based controllers can achieve better robustness against uncertainties of environments through sim-to-real learning. However, the corresponding learned gaits are in general overly conservative and unatural. In this paper, we propose a new framework for learning robust, agile and natural legged locomotion skills over challenging terrain. We incorporate an adversarial training branch based on real animal locomotion data upon a teacher-student training pipeline for robust sim-to-real transfer. Empirical results on both simulation and real world of a quadruped robot demonstrate that our proposed algorithm enables robustly traversing challenging terrains such as stairs, rocky ground and slippery floor with only proprioceptive perception. Meanwhile, using diverse gait patterns, the gaits are more agile, natural, and energy efficient compared to the baselines. Both qualitative and quantitative results are presented in this paper. Project page and videos in \url{https://sites.google.com/view/adaptive-multiskill-locomotion}.

\end{abstract}

\section{INTRODUCTION}
Locomotion controller design for legged robots has been an important research topic in robotics. For decades, researchers have developed successful model-based control approaches to control legged robots \cite{di2018dynamic,winkler2018gait,byl2009dynamically}. However, these controllers generally involve extensive human engineering, and are difficult dealing with complex environments such as challenging terrains. Recently, reinforcement learning becomes an increasingly popular control approach for legged robots. A general paradigm involves training control policies in simulation to optimize expected cumulative rewards, followed by transferring these policies to real robots \cite{rudin2022learning}. In comparison to traditional model-based control approaches, reinforcement learning reduces the need for extensive human engineering and has shown more impressive results, including traversing challenging terrains \cite{lee2020learning, kumar2021rma}, recovering from unhealthy states \cite{lee2019robust}, and navigating complex environments \cite{yang2021learning}.

\begin{figure}[htbp]
\centering     
\begin{minipage}{0.5\textwidth}
        \begin{minipage}[t]{0.25\textwidth}
            \centering
            \includegraphics[width=4cm]{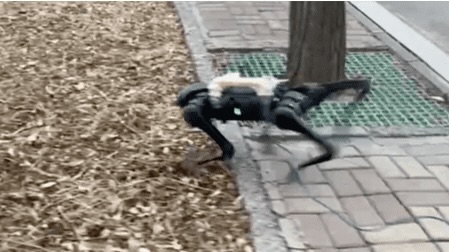}\vspace{0.1cm}
            \includegraphics[width=4cm]{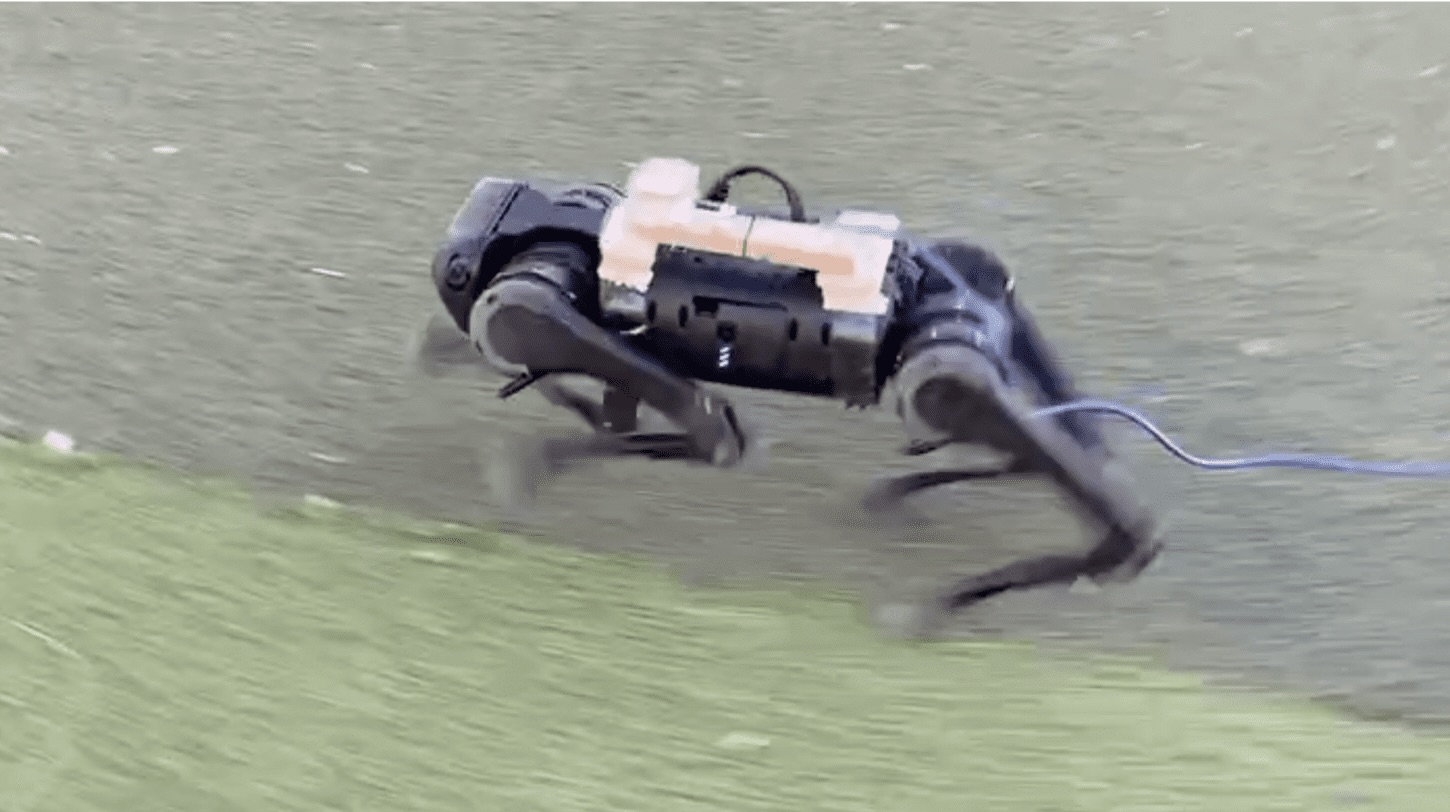}
        \end{minipage}%
        \hspace{2cm}
        \begin{minipage}[t]{0.25\textwidth}
            \centering
            \includegraphics[width=4cm]{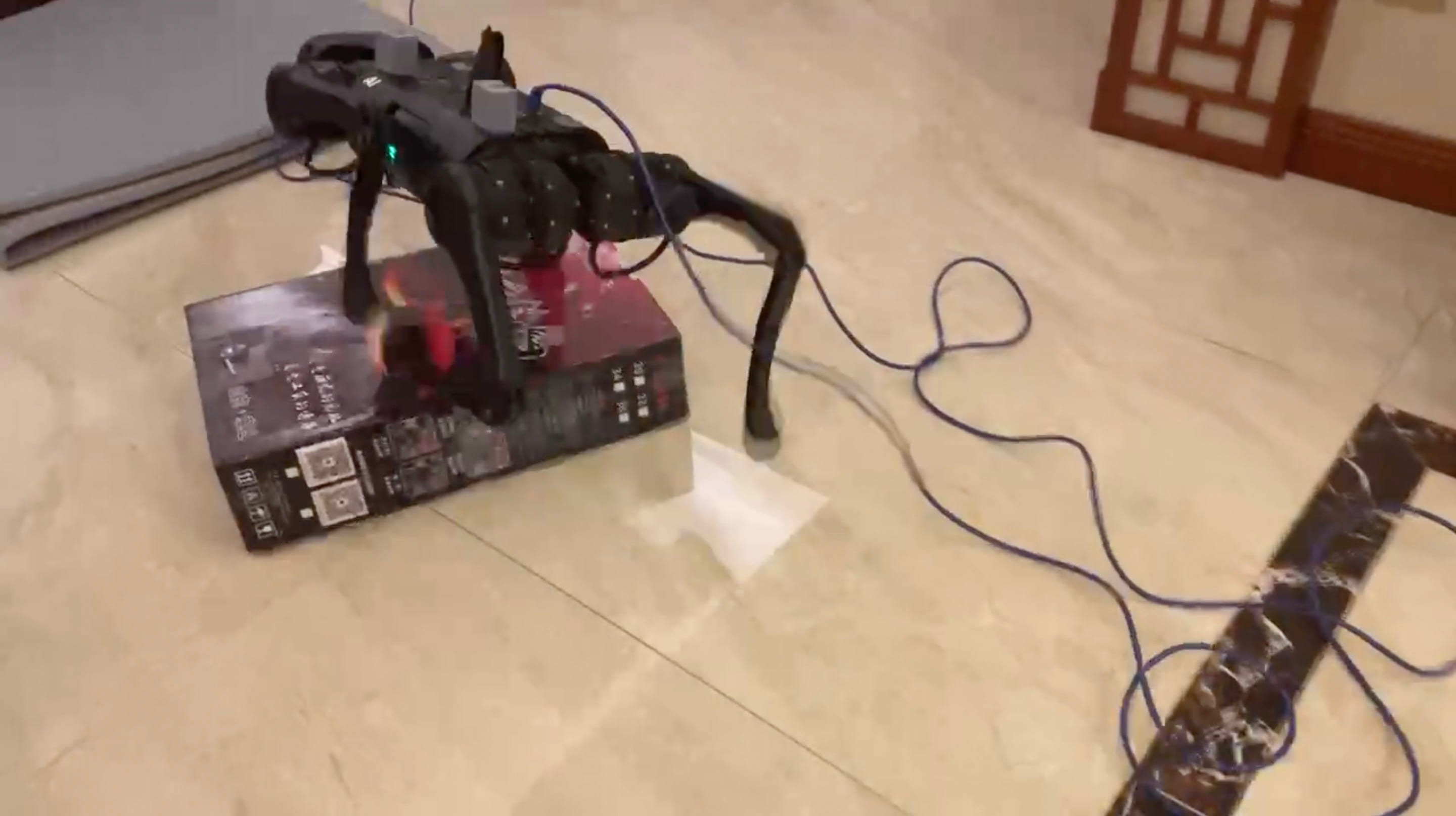}\vspace{0.1cm}
            \includegraphics[width=4cm]{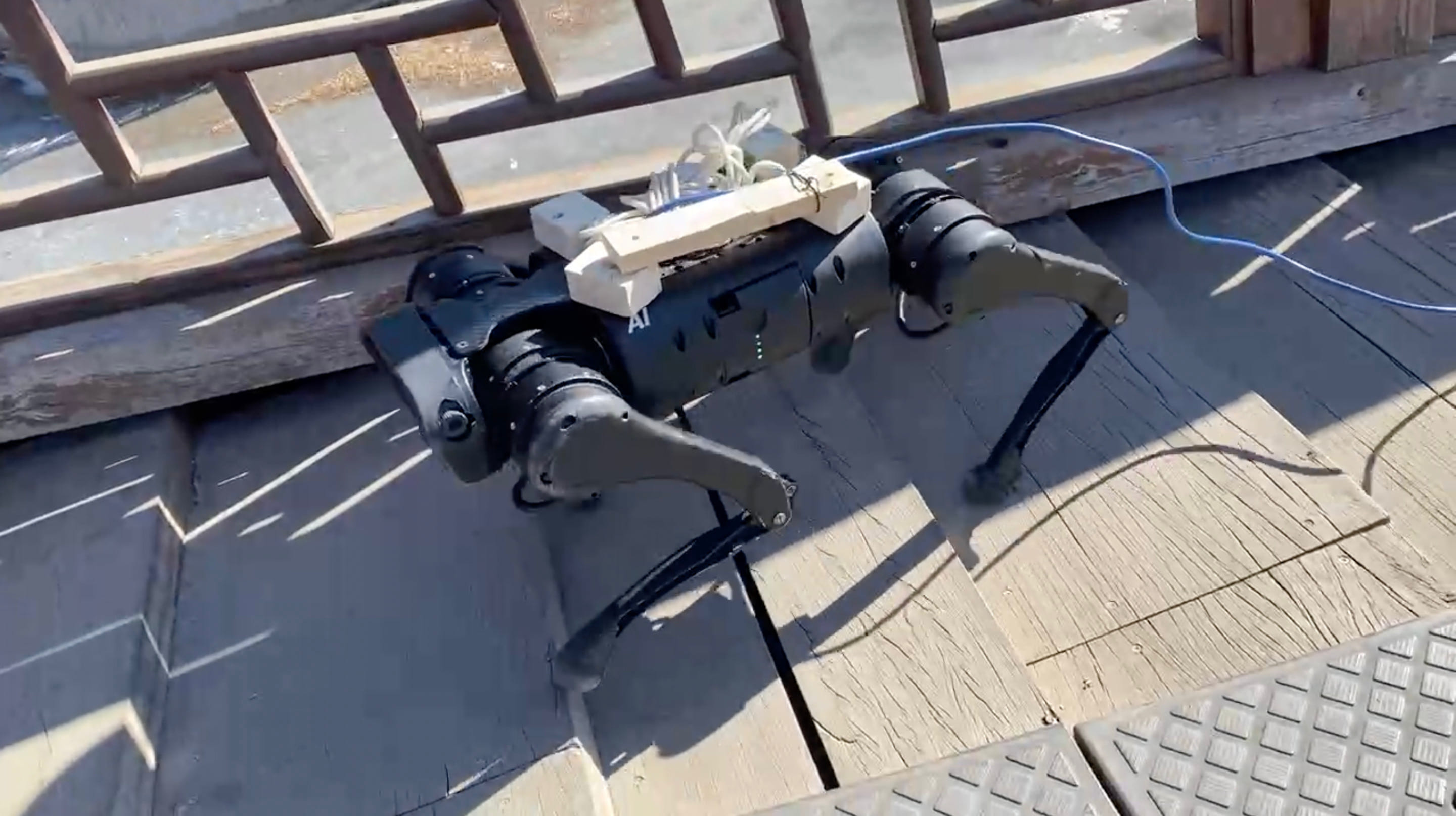}
        \end{minipage}
\end{minipage}
\caption{Experiments in real world.}
\label{real_terrain}
\vspace{-0.5cm} 
\end{figure}

While sim-to-real reinforcement learning exhibits robust legged locomotion skills with appealing properties, in practice, directly optimizing a task reward can lead to policies that produce behaviors undesirable to be applied in real robots, such as unnatural gaits, large contact forces, and high energy consumption. To address these challenges, previous studies have primarily employed intricate reward functions that penalize undesirable behaviors while promoting specific gait patterns\cite{rudin2022learning}. Nevertheless, the process of reward engineering is laborious, and the resulting gaits still frequently appear unnatural.

To address the challenges posed by reward engineering and to achieve more natural gaits, adversarial motion priors (AMP) \cite{peng2021amp} a promising approach which leverages motion capture data and utilizes adversarial imitation learning to acquire locomotion tasks that closely resemble real-world motion data. While such method has demonstrated successful transfer from simulation to a real quadrupedal robot \cite{escontrela2022adversarial}, the learned control policy is limited to traversing flat terrain in a laboratory environment, thereby lacking the capability to handle challenging terrains such as stairs or slippery ground. An intuitive extension is to train the control policy in simulation environments that incorporate different types of terrain. However, based on our experiment results, policies trained with this approach fail to achieve satisfied rewards even within simulation.

In this paper, we propose a new framework which enables learning not only robust, but also agile and natural legged locomotion skills over challenging terrains in the wild. We adopt a teacher-student training paradigm to enable adaptation in real world, where the teacher policy is trained with encoded privileged information, and the student policy is trained to infer the information from historical observations. Several techniques are applied to enhance the robustness of the policy. We then incorporate an adversarial training branch into this teacher-student training pipeline to match the learned skills with real animal motion capture data. We evaluate our method on a quadruped robot in both simulation and real world. Experiment results show that our method successfully learn legged locomotion skills to traverse challenging terrains such as stairs, rocky ground and slippery floor. Using diverse gait patterns, the learned gaits are more agile, natural, and energy efficient compared to baselines,. We also find the policy is able to smoothly transit different gaits for different velocity command and terrains. In summary, our contributions are:
\begin{itemize}
\item Present a framework that empower the robot with high level of robustness and naturalness to move in the wild while exhibiting diverse gait patterns.
\item To the best of our understanding, this is the first learning-based method enabling quadrupedal robots to gallop in the wild.
\end{itemize}

\section{RELATED WORK}
\subsection{Reinforcement Learning for Legged Locomotion}
Recent advancements in deep reinforcement learning for legged locomotion have demonstrated its promising future. Lee et al. \cite{lee2020learning} applied teacher-student training to the quadruped robot ANYmal, resulting in a robust controller capable of traversing challenging terrains, which is similar to the teacher-student training paradigm as ours. Peng et al. \cite{peng2020learning} introduced the use of Deep Mimic \cite{peng2018deepmimic} to learn robotic locomotion skills by imitating animals. We adopted the similar motion retargeting technique as \cite{peng2020learning}. Similar to \cite{lee2020learning}, Kumar et al. \cite{kumar2021rma} trained locomotion policies with rapid motor adaptation, enabling them to quickly adapt to environmental changes. Building upon this, Kumar et al. \cite{kumar2022adapting} extended the RMA algorithm to the bipedal robot Cassie. Yang et al. \cite{yang2021learning} employed a cross-modal transformer to learn an end-to-end controller for quadrupedal navigation in complex environments. Ji et al. \cite{ji2022concurrent} trained a neural network state estimator to estimate robot states that cannot be directly inferred from sensory data. Escontrela et al. \cite{escontrela2022adversarial} utilized Adversarial Motion Priors (AMP) to train control policies for a quadrupedal robot, highlighting that AMP can effectively substitute complex reward functions. The relationship between \cite{escontrela2022adversarial} and ours is that ours further adapted the AMP algorithm to work on challenging terrains. Sharma et al. \cite{sharma2020emergent} trained a reinforcement learning controller using unsupervised skill discovery and successfully transferred it to a real quadruped robot. Xie et al. \cite{xie2021dynamics} revisited the necessity of dynamics randomization in legged locomotion and provided suggestions on when and how to employ dynamics randomization. Bohez et al. \cite{bohez2022imitate} trained a low-level locomotion controller for a quadruped robot by imitating real animal data, utilizing this controller to accomplish various tasks. Margolis et al. \cite{margolis2021learning} trained policies to perform jumps from pixel inputs, while Miki et al. \cite{miki2022learning} trained a locomotion controller using observations of the height map of the terrain around the robot's base. Rudin et al. \cite{rudin2022learning} employed massively parallel simulation environments to significantly accelerate the training process of a locomotion controller. Margolis et al. \cite{margolis2022rapid} trained a locomotion controller for the Mini Cheetah robot, enabling it to achieve speeds of up to 3.9m/s, surpassing traditional controllers' speeds by a large margin. Other notable works include directly learning locomotion skills in the real world \cite{wu2022daydreamer, smith2022walk}, as well as learning locomotion skills for bipedal robots \cite{xie2019iterative, siekmann2021blind, siekmann2021sim}.

\subsection{Motion Control from Real World Motion Data}
Imitating a reference motion dataset offers an approach to designing controllers for skills that are challenging to manually encode. Pollard et al. \cite{pollard2002adapting,grimes2006dynamic,suleiman2008human,yamane2010controlling} employ motion tracking techniques, where characters explicitly mimic the sequence of poses from reference trajectories. Learning from real-world motion provides an alternative to crafting complex rewards for synthesizing natural motion. Peng et al. \cite{peng2018deepmimic} adapt reinforcement learning (RL) methods to learn robust control policies capable of imitating a wide range of example motion clips. Leveraging GAN-style training, Peng et al. \cite{peng2021amp} learn a "style" reward from a reference motion dataset to control the character's low-level movements, while allowing users to specify high-level task objectives. Escontrela et al. \cite{escontrela2022adversarial} utilize the framework proposed by Peng et al. \cite{peng2021amp} to train a locomotion policy for a quadrupedal robot to traverse flat ground. Additionally, Peng et al. \cite{peng2022ase} present a scalable adversarial imitation learning framework that enables physically simulated characters to acquire a wide repertoire of motor skills, which can be subsequently utilized to perform various downstream tasks.
\section{METHOD}

\begin{figure*}[htbp]

\begin{minipage}[t]{0.6\textwidth}
\centering
\includegraphics[width=\textwidth]{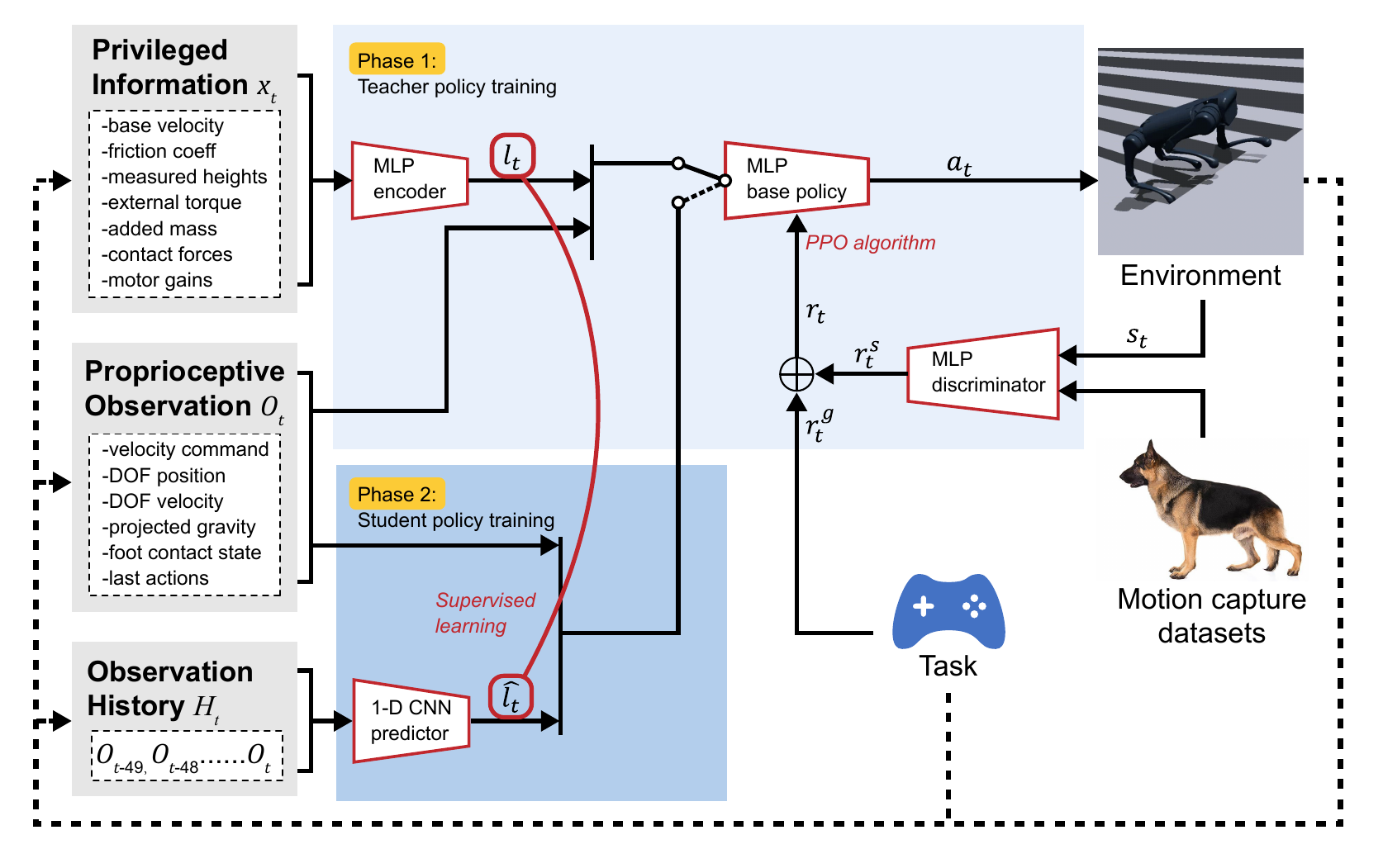}
\caption{Overview of the traning and control framework.}
\label{overview}
\end{minipage}
\hfill
\begin{minipage}[t]{0.4\textwidth}
\centering
    \includegraphics[width=\textwidth]{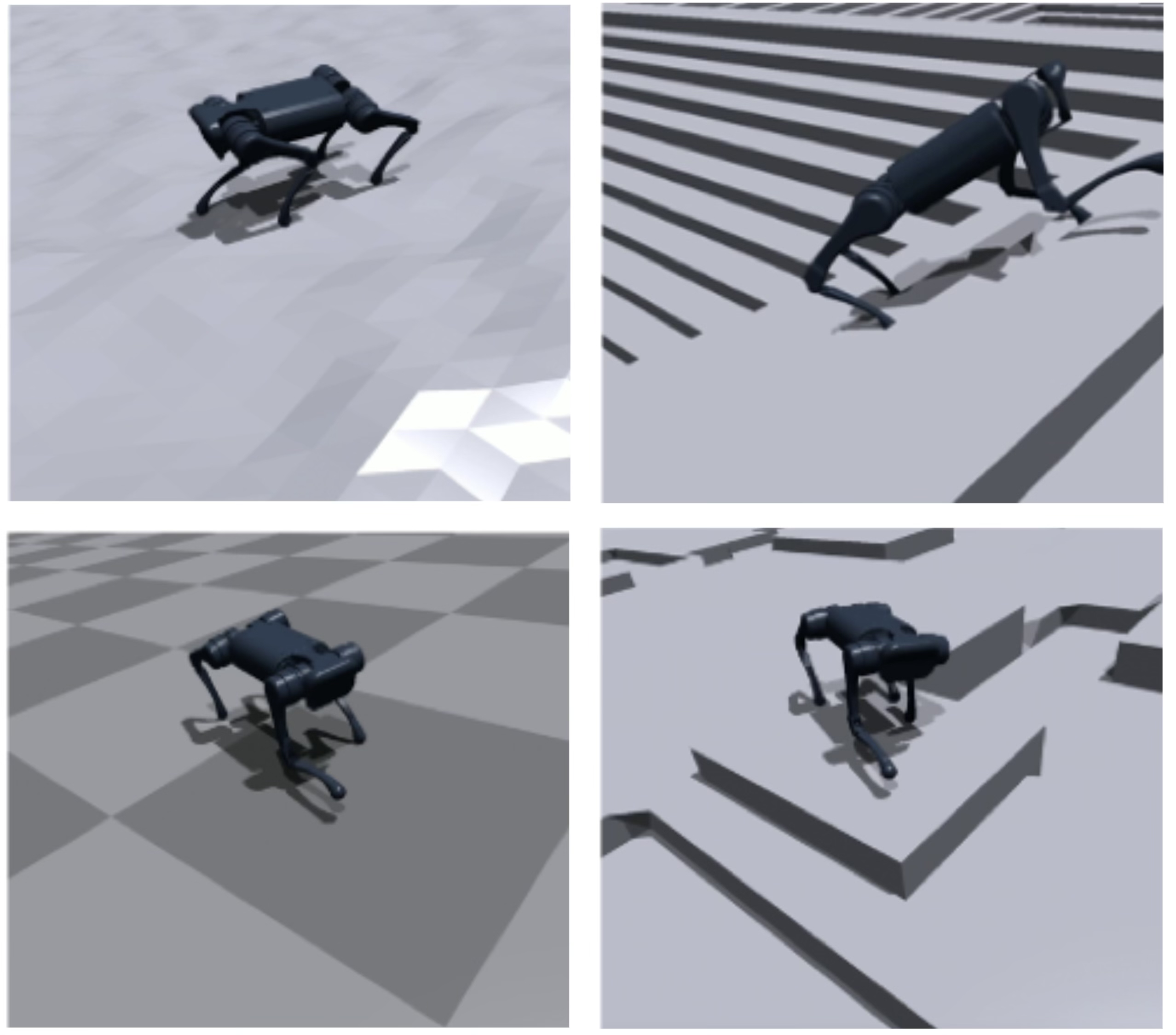}
    \caption{Terrains in simulation. }
    \label{simterrains}
    \end{minipage}
    \vspace{-0.5cm}
\end{figure*}

The proposed approach comprises several building blocks which mainly support robust sim-to-real learning as well as natural gait learning from motion capture reference. An overview of the proposed framework is shown in Figure \ref{overview}. We first have a phase 1 training process, which learns a teacher policy using both proprioceptive observation and the privileged information. An adversarial training process is running simultaneously to enforce agile and natural gait from motion capture reference data. Then at the phase 2 training process, we learn a student policy which takes the historical proprioceptive observations and output the final actions with the policy. This policy are directly deployed to the real robot which bridges the sim-to-real gap. In this section, we will introduce the details of each component.

\subsection{Robust Sim-to-Real Locomotion Learning}
To enable robust legged locomotion through sim-to-real learning, we adopt a teacher-student training framework, augmented with several tricks to enhance sim-to-real transfer.
\subsubsection{Teacher-Student Training Framework}
Inspired by previous works for robust legged locomotion learning ~\cite{lee2020learning, kumar2021rma}, we integrate the teacher-student training paradigm into our framework. The teacher policy encodes privileged information of the environments and the robot from the simulation, while the student policy only takes observations directly available from sensors on the real robot. Thus, the privileged simulation information enables good performance for the teacher policy in challenging environments, which is then adopted to teach the student policy to perform equally well in the real world.


\textbf{Teacher Policy Training}: 
We can formulate the teacher policy training problem as a Markov Decision Process (MDP)~\cite{puterman1990markov}. An MDP can be represented by a tuple $(\mathcal{S}, \mathcal{A}, \mathcal{P}, R, \gamma)$, where $\mathcal{S}$ and $\mathcal{A}$ denote the state and action spaces, respectively. The transition probability function $\mathcal{P}: \mathcal{S}\times\mathcal{A}\times\mathcal{S}\rightarrow [0, 1]$ maps the current state and action to the probability distribution of the next state, based on the robot and environment dynamics. The reward function $R: \mathcal{S}\times\mathcal{A}\rightarrow\mathbb{R}$ assigns a real value to each state-action pair. The discount factor $\gamma$ determines the importance of future rewards. The goal of reinforcement learning is to find a policy $\pi:\mathcal{S}\rightarrow\mathcal{A}$ to maximize the the expected future cumulative rewards $J_R(\pi)=\mathbb{E}_{\tau\sim\pi}\left[\sum_{t=0}^{\infty}\gamma^t R(s_t, a_t)\right]$.

In our work, the state $s$ is composed of both the proprioceptive observation $O_t$ and a latent vector $l_t$. Here $l_t$ contains encoded privileged information using an encoder $l_t=\mu(x_t)$. Then a base policy $\pi$ maps the concatenated state $s_t=(l_t, O_t)$ to the action command $a_t$. Compared to previous works, we incorporate additional data in the privileged information, such as the robot base velocity and measured terrain heights represented as 3D points. By learning the encoder along with the reinforcement learning process, the encoded latent $l_t$ is expected to contain rich information about the robot dynamics, state estimation, and environment factors. Detailed definition about the privileged information, the proprioceptive observation, as well as the action space can be found in the project website. We choose proximal policy optimization (PPO) as our backbone reinforcement learning algorithm. Detailed design of the network architectures and algorithm hyper-parameters can be found in the project website.

\textbf{Student Policy Training}: 
The teacher policy cannot be directly deployed to the real robot, since the privileged information is generally only available in simulation, but non-trivial or impractical to obtain in real world. Therefore, after training the teacher policy, we train a student policy which mimic the functionalities of the teacher policy, but is feasible to be directly deployed on the real robot. To do this, we train another encoder $\hat{\mu}$ (named 'predictor'), which takes a series of historical proprioceptive observations ${O_{t-T}, ..., O_{t-1}, O_t}$ as inputs, instead of the privileged information. Such historical observations are easily obtained on real robot by creating memory buffer for sensor inputs. The predictor is trained using supervised learning to minimize the error between the predictor output $\hat{l}_t$ and the ground truth latent $l_t$: $||\hat{l}_t-l_t||^2$. The intuition underneath is that, according to the general assumption of partially observable Markov decision process (POMDP) \cite{kaelbling1998planning}, the unobservable true state such as the privileged information can be recovered from historical observations. After obtaining the latent $\hat{l}_t$, we use the same base policy $\pi$ with the teacher policy to compute the action $a_t$, which is then directly applied to the real robot actuators. Detailed implementation of this part can be found in the project website.



\subsubsection{Enhancing Sim-to-Real Transfer}
Upon the teacher-student training paradigm, we also incorporate several important techniques to enhance the sim-to-real transfer performance, including domain randomization, terrain curriculum, and action filtering.

\textbf{Domain Randomization}: Domain randomization is the most important practical techniques for robot sim-to-real transfer. In general, the sim-to-real problem mainly lies on the gap between the dynamics models of simulation and real world, as well as the the gap between their state distributions. In essence, if we randomize both the dynamics and the state distribution in the simulation, so that the randomization range covers the real world setting, then learned policy should be robust in real world environments. We randomize both robot dynamics, such as mass and motor gain, as well as environment dynamics, such as friction coefficient. Besides randomizing dynamics, we also add perturbations to the robot on its linear velocity, angular velocity, as well as actuator torques. Furthermore, we add noises to sensor observations. Please refer to the project website for detailed ranges of the randomization and perturbation we apply.



\textbf{Terrain Curriculum}: Besides the above randomization terms, we generate random terrains in simulation during training, in order to generalize to various ground environment in the real world. We adopt the game-inspired terrain curriculum proposed in~\cite{rudin2022learning}. Specifically, we utilize four types of terrains: plane ground, uniform noise, discrete obstacles, and stairs. Before proceeding to a more challenging type of terrain, the robot needs to successfully traverse the current terrain and achieve a satisfied task reward. The threshold we use to increase terrain difficulty consists: (1)The robot successfully crosses the center of a terrain block within a single episode; (2)The linear velocity tracking reward surpasses $80\%$  of the maximum achievable reward which corresponds to 'perfectly' accurate tracking.

In contrast, the robots are reset to easier terrains if they fail to travel more than half of the distance required by their command linear velocity within an episode. This adaptive curriculum mechanism enables us to stably learn robust locomotion skills for the robot.

\textbf{Action Filtering}: 
We observed that applying a low-pass filter to the output actions can significantly improve the smoothness of motions, enabling better sim-to-real transfer. The filter is defined as:
$$
u_{t}=0.2*u_{t-1}+0.8*a_{t}
$$
where $u_{t}$ represents the filtered target joint angles applied to the low-level PD controllers, and $a_{t}$ corresponds to the action output by the neural network policy.

\subsection{Natural Gait Learning with Motion Capture Reference}
We hope the learned locomotion skills to be not only robust, but also natural and agile just like real animals. Inspired by adversarial motion priors (AMP) \cite{peng2021amp}, we incorporate an adversarial motion style matching process into our framework, in order to learn robust, agile, and natural legged locomotion skills.

\textbf{Adversarial Motion Style Matching}:
In order to learn agile and natural gaits, our designed reward for the reinforcement learning problem consists of both a "task" reward $r^g_t$ and a "style" reward $r^s_t$.\ The overall reward function is given by $r_t=\omega^gr^g_t+\omega^sr^s_t$. The ratio of these two part is quite critical to the robot's performance. In this work, we chose $\omega^g$ to be 0.35, while $\omega^s = 0.65$. We have provided a comprehensive explanation for this choice in the Appendix C.
  The task reward is defined based on the specific task we aim to accomplish, here it consists of a linear velocity command tracking reward and an angular velocity command tracking reward:
\begin{equation}
\label{rew_tracking}
r_t^g=w^v \exp \left(-\left|\hat{v}_t^{\mathrm{xy}}-v_t^{\mathrm{xy}}\right|\right)+w^\omega \exp \left(-\left|\hat{\omega}_t^z-\omega_t^z\right|\right)
\end{equation}
where $w^v$, $w^\omega$, and $w^\tau$ are the coefficients. $\hat{\vec{v}}_t^{\mathrm{xy}}$ and $\hat{\omega}_t^z$ represent the linear and angular velocity commands, respectively. To ensure robustness and learn diverse gait patterns, different ranges of velocity commands are defined for each terrain type, as listed in the website. The velocity commands are randomly sampled from the specified ranges.

The style reward is generated by a discriminator $D_\phi$, which is trained to classify whether the given state transition samples are from the motion capture dataset or from the policy rollouts, where $\phi$ denotes the discriminator's parameters. The optimization objective of the discriminator is as follows:
\begin{equation}
\label{disc}
\begin{aligned}
\underset{\phi}{\arg \min}\; & \mathbb{E}_{\left(s, s^{\prime}\right) \sim \mathcal{D}}\left[\left(D_\phi\left(s, s^{\prime}\right)-1\right)^2\right] \\
& +\mathbb{E}_{\left(s, s^{\prime}\right) \sim \pi_\theta(s, a)}\left[\left(D_\phi\left(s, s^{\prime}\right)+1\right)^2\right] \\
& +\frac{w^{\mathrm{gp}}}{2} \mathbb{E}_{\left(s, s^{\prime}\right) \sim \mathcal{D}}\left[\left\|\nabla_\phi D_\phi\left(s, s^{\prime}\right)\right\|^2\right],
\end{aligned}
\end{equation}
where $\mathcal{D}$ denotes the motion capture dataset, The first two terms incentivize the descriminator to output 1 for transition pairs from the mo-cap dataset, while output -1 for transition pairs from the policy rollouts. $\omega^{gp}$ is the coefficient for gradient penalty which reduces oscillations in the adversarial training process. The style reward is then defined as:
\begin{equation}
    r_t^s(s_t, s_{t+1})=\max \left\{0, 1-0.25(D_\phi(s_t, s_{t+1})-1)^2\right\}
\end{equation}

Therefore, the policy is trained through reinforcement learning to maximize the reward function as a generator, while the discriminator is trained using both the motion dataset $\mathcal{D}$ and the data generated during policy rollouts, forming an adversarial motion style matching framework. 

\textbf{Motion Capture Data Reference}:
We utilize high-quality dog motion capture data provided by Zhang et al. \cite{zhang2018mode}. The dataset was obtained from a real German shepherd, which exhibits different morphology compared to our quadrupedal robot. To adapt the dog motion data to our robot, we apply inverse kinematics for motion retargeting as employed in Peng et al. \cite{peng2020learning}. Furthermore, we enhance the motion capture data by mirroring the dataset, specifically by switching the joint angle sequences between the left and right legs and reflecting the base position of the robot. Empirically, we find that this is crucial for the sim-to-real transfer of gallop gait.
\section{EXPERIMENTS}
We use Isaac Gym~\cite{makoviychuk2021isaac} simulator for training and use Unitree A1 as our robot platform in both simulation and real world. The robot's motor encoders provide joint angles and angular velocities, the IMU provides projected gravity information, and the foot force sensors obtain binary contact states. we compare the performance of our approach with two baselines: 
\begin{itemize}
\item \textbf{Complex rewards}: Policy trained with typical model-free RL method using complex hand-designed reward function as in~\cite{rudin2022learning}. 
\item \textbf{AMP}: Policy trained using adversarial motion priors as style reward to learn agile and natural legged locomotion skills~\cite{escontrela2022adversarial}.
\end{itemize}

We conduct both simulation and real world experiments to evaluate out method, which demonstrate that our method outperforms baselines by learning robust, agile, natural and energy-efficient legged locomotion skills.

\subsection{Simulation Experiments}\label{sim_exp}
\textbf{Metrics:} In simulation experiments, we compare the performance of our approach with the comparison baselines on following metrics: 
\begin{itemize}
\item \textbf{Success rate}: Success means no falling before reaching the goal position in less than a threshold time. The success rate is calculated as the ratio of successful trials to the total number of experiments conducted.
\item \textbf{TTF}: The time to fall normalized by the maximum duration of trajectories.
\item \textbf{Command tracking accuracy}: The average velocity command tracking reward as defined in Equation~\eqref{rew_tracking}.
\item \textbf{Energy efficiency}: The energy efficiency is estimated by computing the average power of motors, which is defined as: $avgPower=\sum_{\text {motors}}[\tau \dot{\theta}]^{+}$, where $\tau$ is the joint torque, and $\dot{\theta}$ is the motor velocity.

\end{itemize}

Note that we only measure energy efficiency and command tracking accuracy of a policy when its success rate exceeds $60\%$. 

\textbf{Terrains}: 
We evaluate the learned policies on three kinds of challenging terrains: 
\begin{itemize}
    \item \textbf{Stairs}: Continuous stairs in surroundings with step height of 14cm and step width of 31cm.
    \item \textbf{Discrete obstacles}: Randomly place discrete obstacles on the ground with height of 15cm, or pit with depth of 15cm.
    \item \textbf{uniform noise}: Uneven ground generated by adding uniform noise to the terrain heights.
\end{itemize}
Examples of different terrains in simulation are show in Figure~\ref{simterrains}.

\textbf{Results}:
Results of the evaluation metrics in simulations are shown in Table~\ref{big_table}  and Table~\ref{sucandttf}. Each experiment is conducted using three polices trained with distinct random seeds. Each policy is subjected to 1000 independent experiments, the average value along with a $95\%$ confidence interval is reported. AMP fails to traverse stairs and discrete obstacles, while Complex Rewards fails to traverse discrete obstacles, so they are omitted in the table. 

We can see that our controller can successfully traverse through a greater variety of complex terrains with higher success rate, this might be due to the diverse motion capture data that enables the robot to switch to the most suitable gait or blend different gaits at different terrains and speeds. Meanwhile, the teacher-student training architecture plays an important role in state estimation and system identification. 

\begin{figure*}
\hsize=\textwidth
        \centering
        \begin{minipage}{0.12\textwidth}
            \includegraphics[width=1\linewidth]{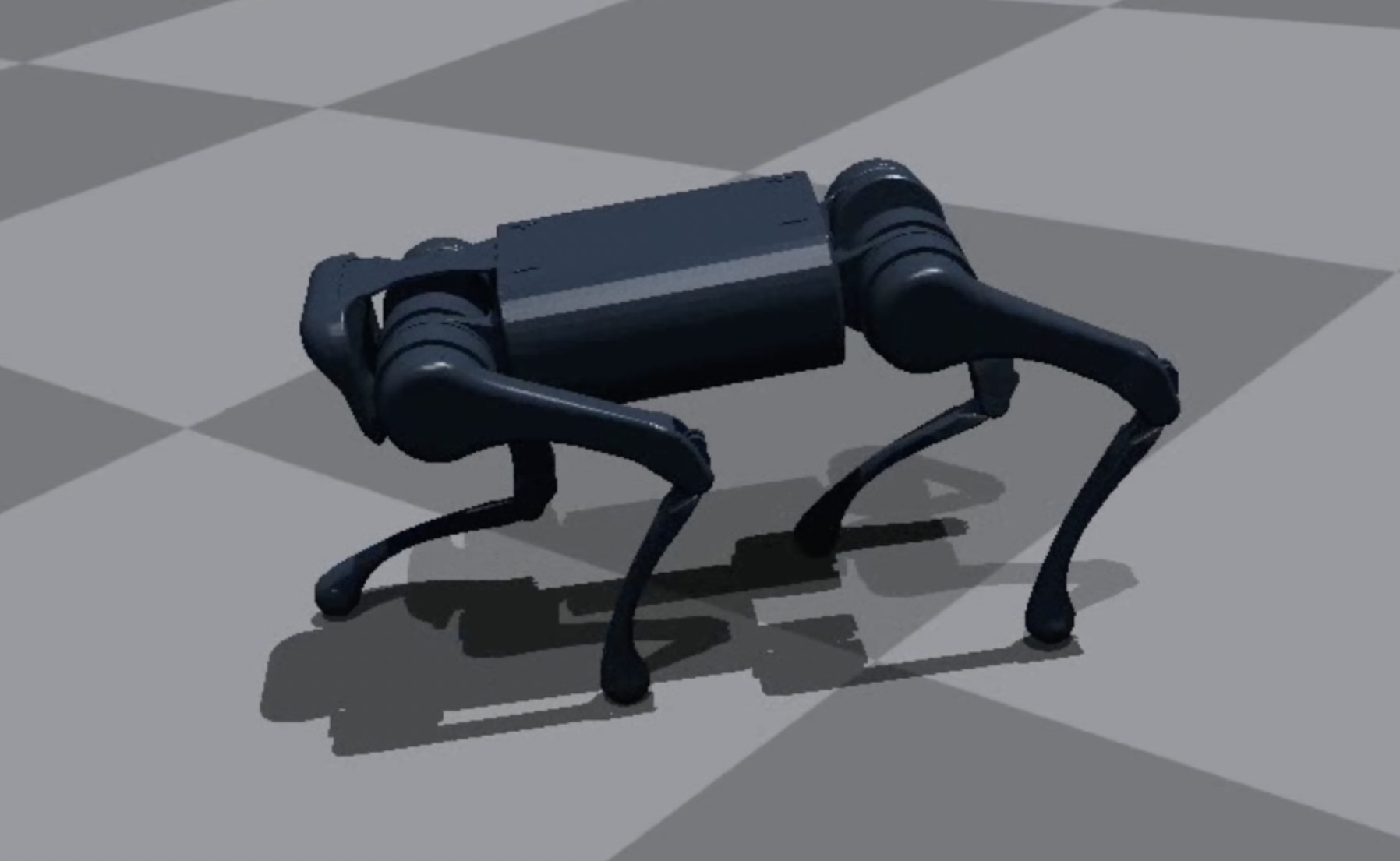}
        \end{minipage}
        \hspace{-6pt}
        \begin{minipage}{0.12\textwidth}
            \includegraphics[width=1\linewidth]{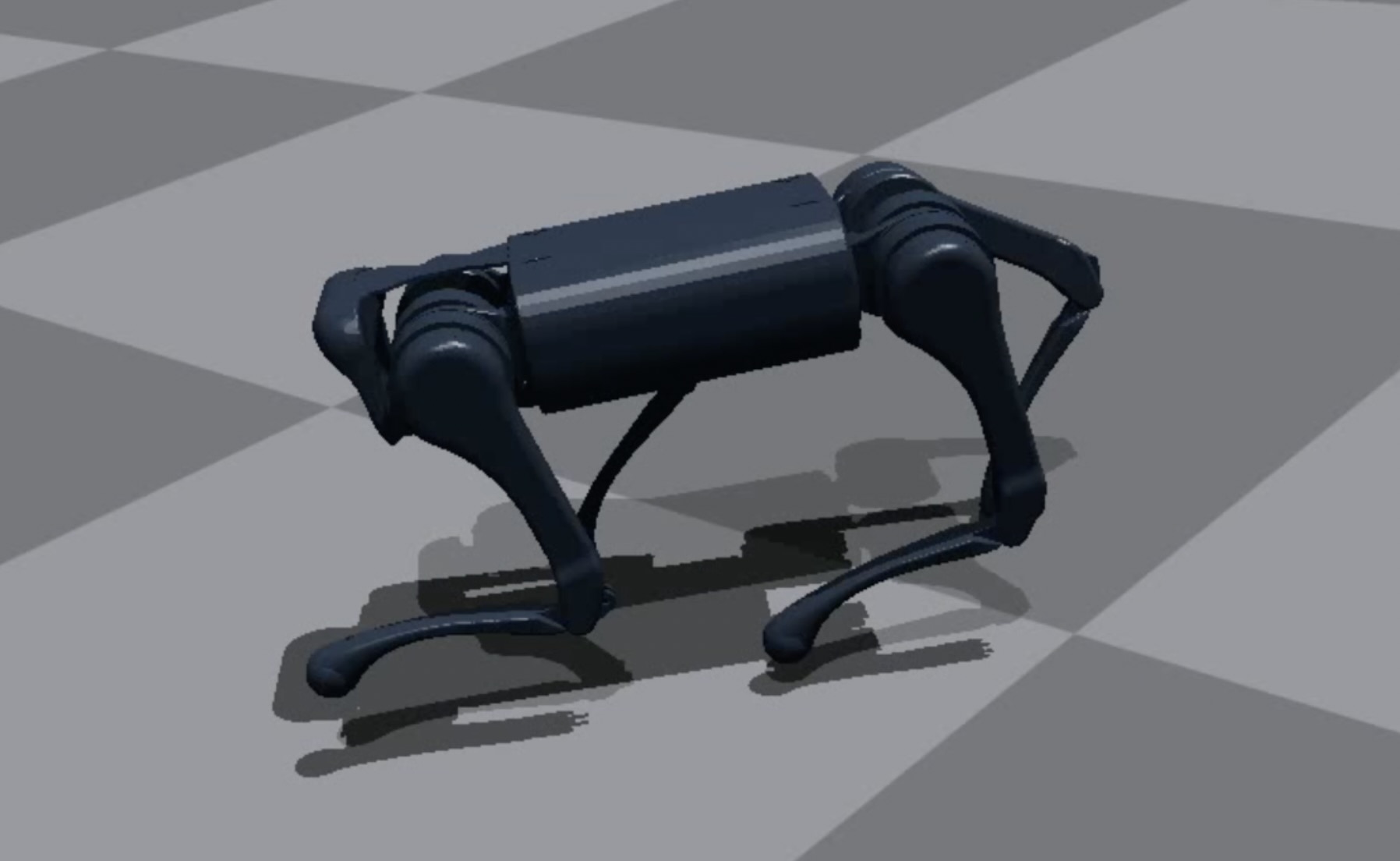}
        \end{minipage}
        \hspace{-6pt}
        \begin{minipage}{0.12\textwidth}
            \centering
            \includegraphics[width=1\linewidth]{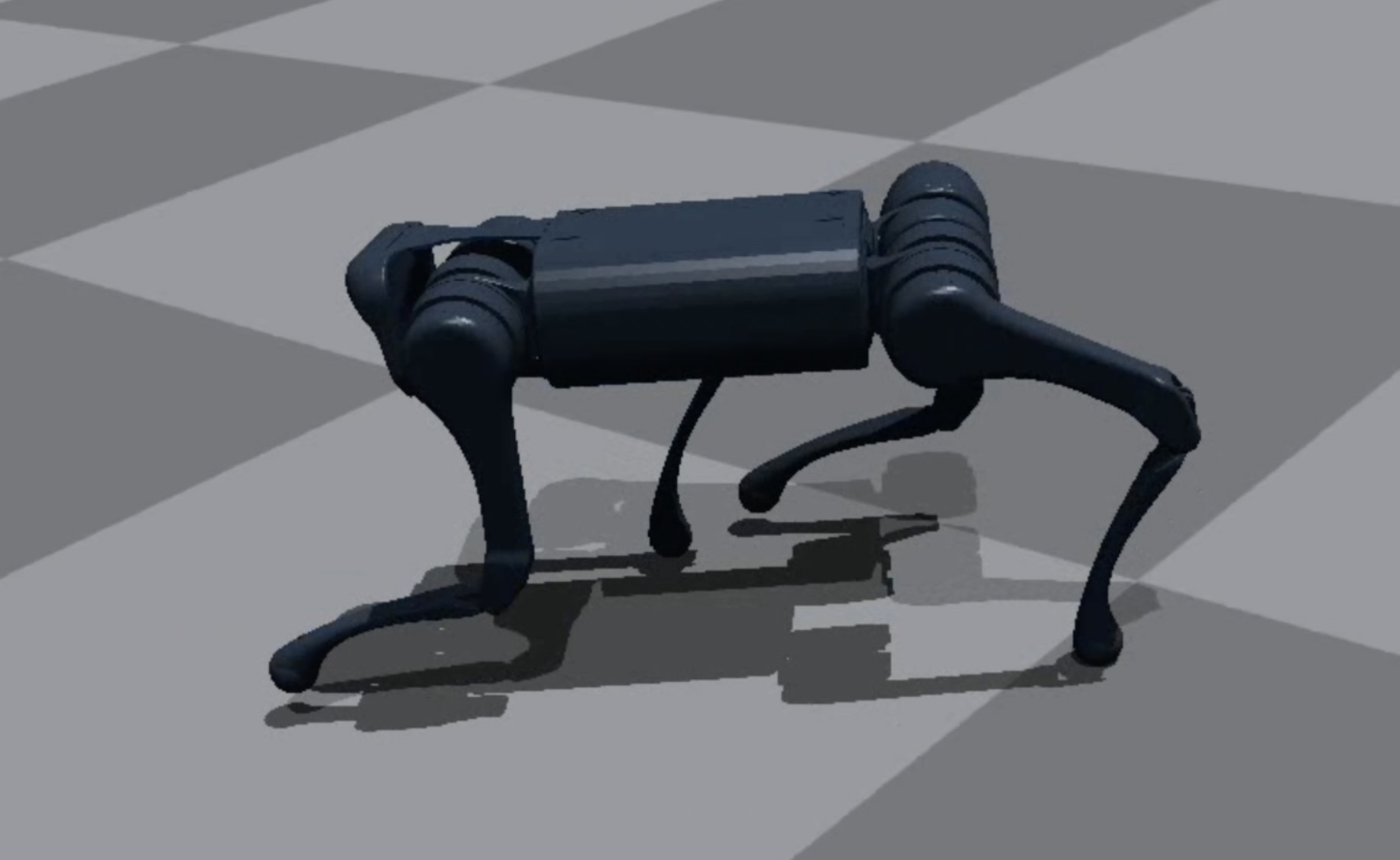}
        \end{minipage}
        \hspace{-6pt}
        \begin{minipage}{0.12\textwidth}
            \centering
            \includegraphics[width=1\linewidth]{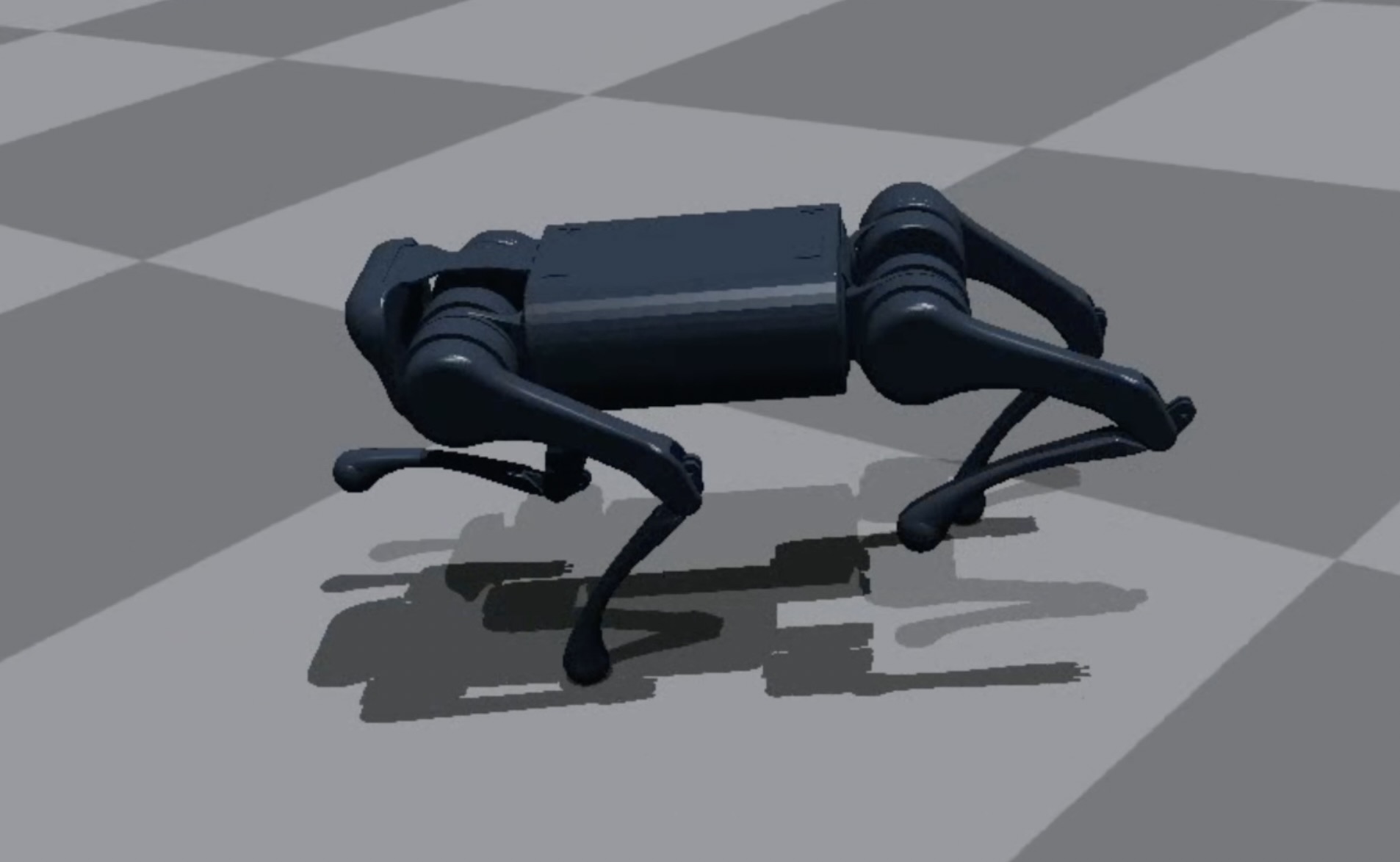}
        \end{minipage}
        \hspace{-6pt}
        \begin{minipage}{0.12\textwidth}
            \includegraphics[width=1\linewidth]{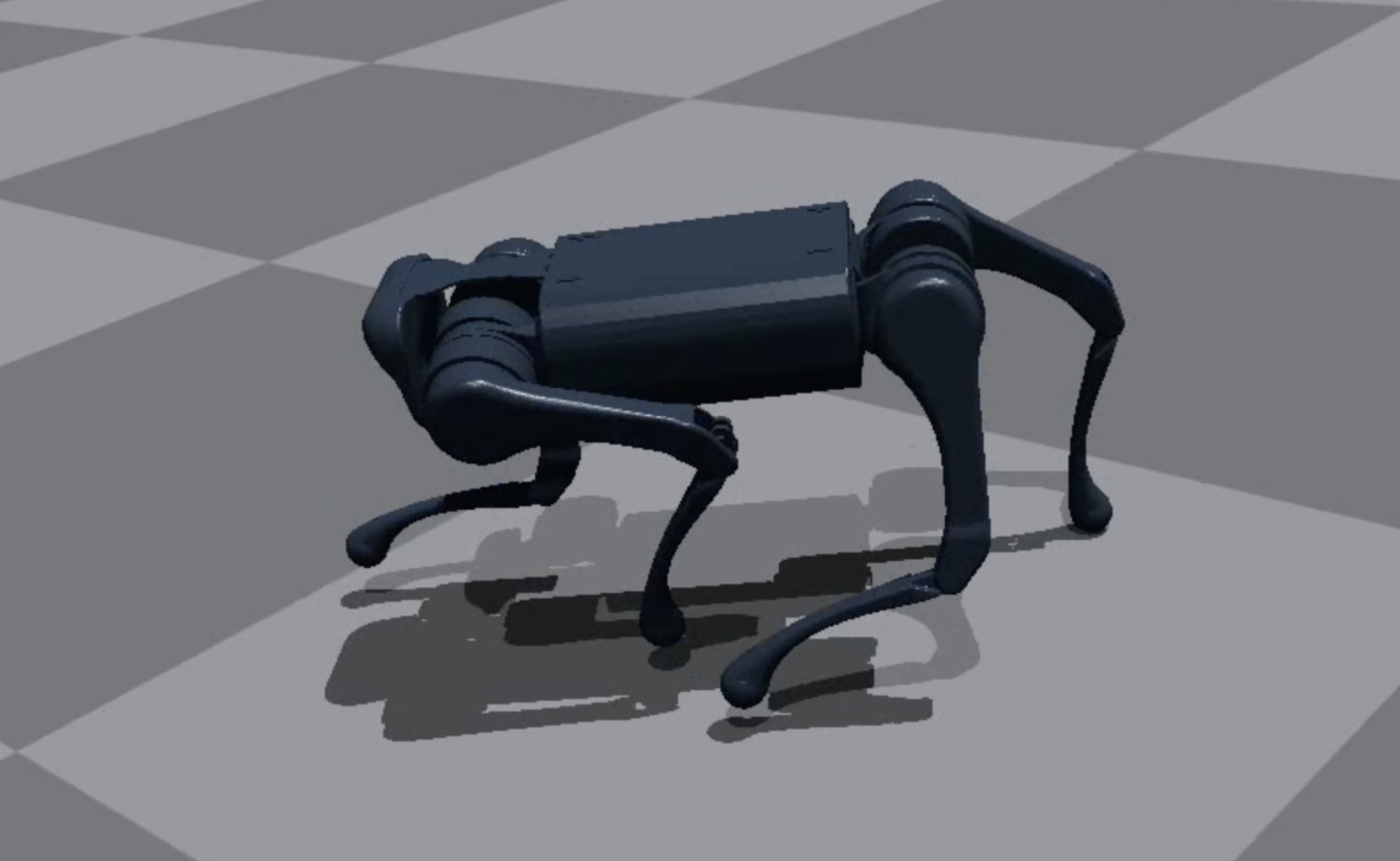}
        \end{minipage}
        \hspace{-6pt}
        \begin{minipage}{0.12\textwidth}
            \includegraphics[width=1\linewidth]{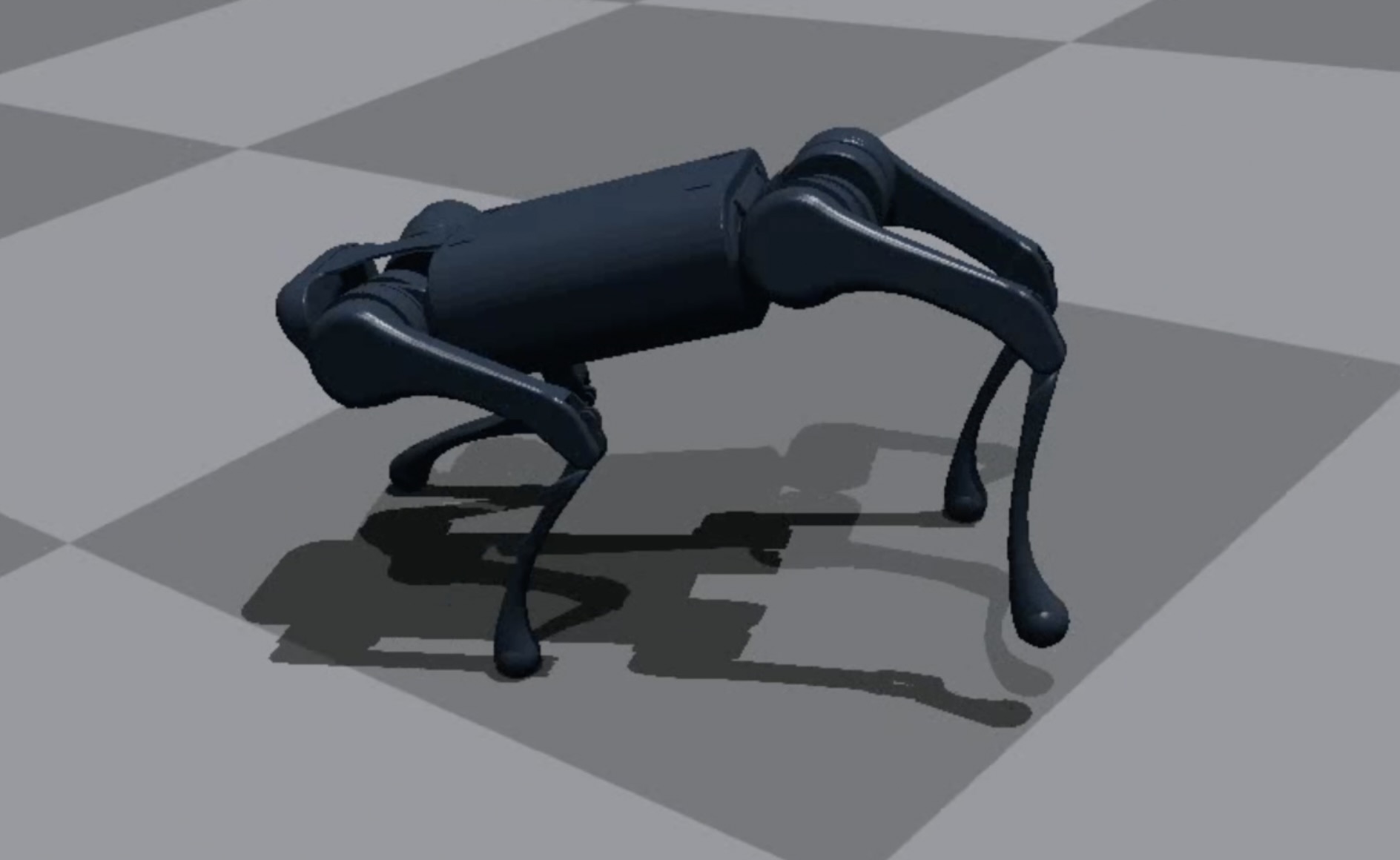}
        \end{minipage}
        \hspace{-6pt}
        \begin{minipage}{0.12\textwidth}
            \centering
            \includegraphics[width=1\linewidth]{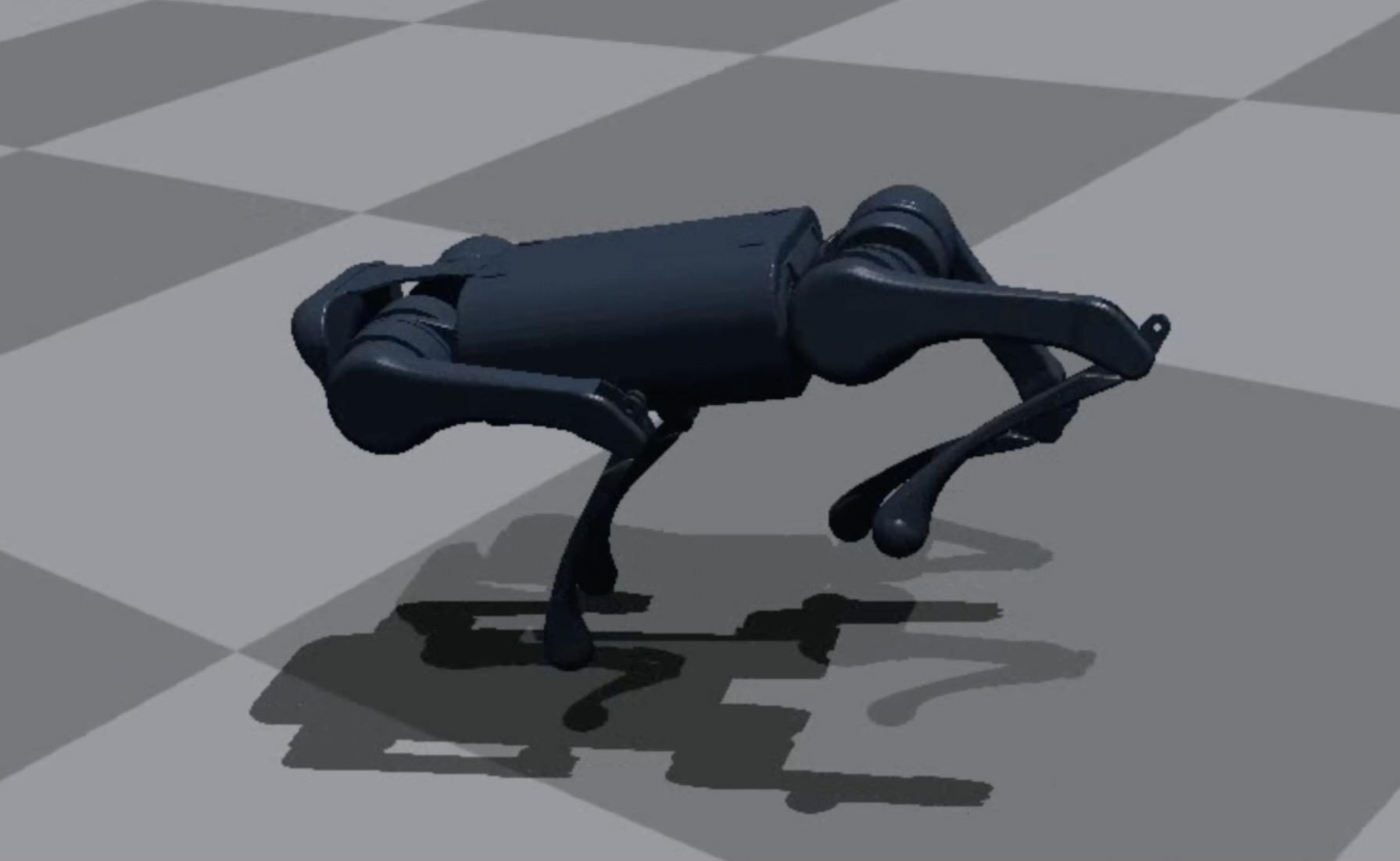}
        \end{minipage}
        \hspace{-6pt}
        \begin{minipage}{0.12\textwidth}
            \centering
            \includegraphics[width=1\linewidth]{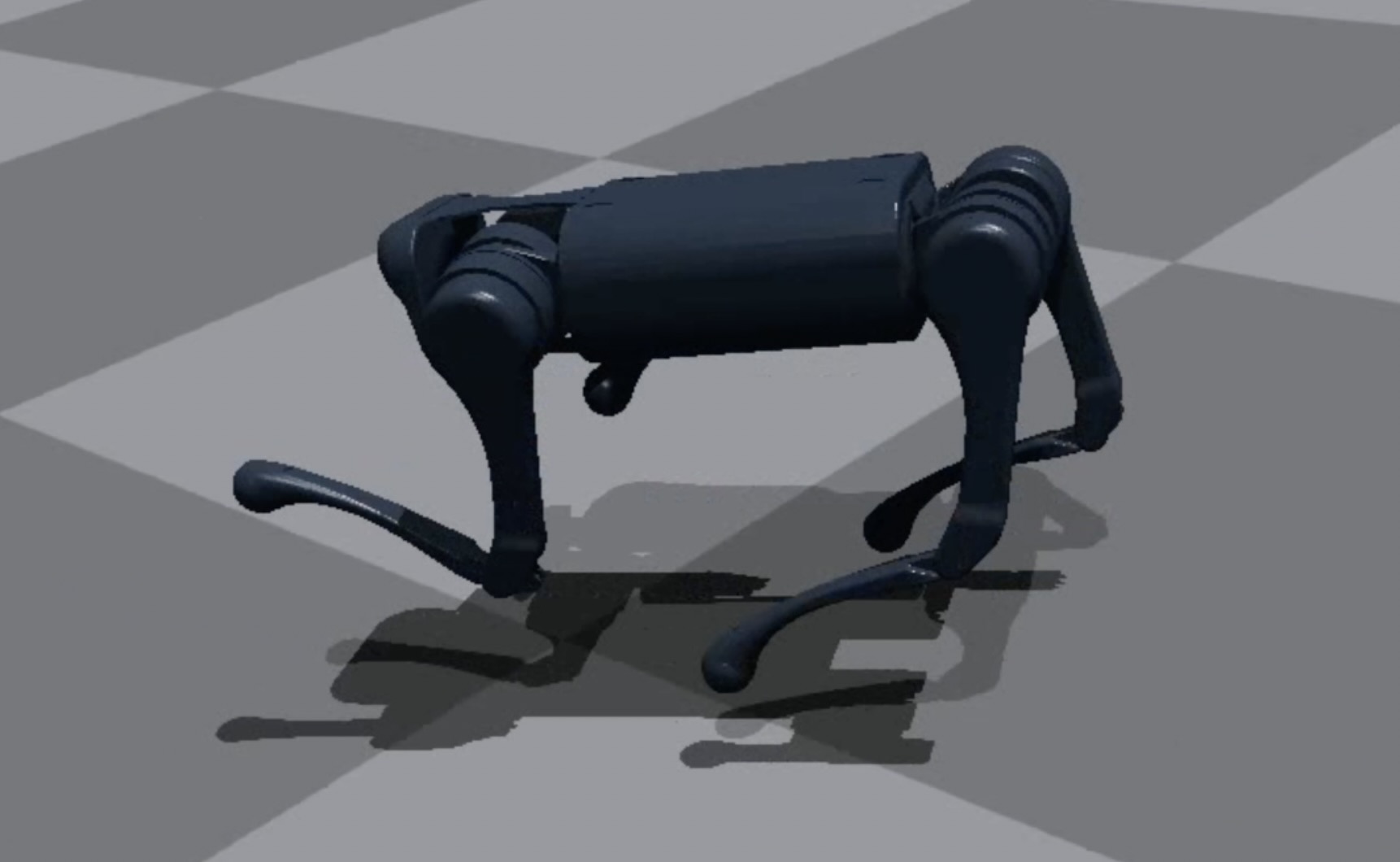}
        \end{minipage}
    \caption{Transition from the 'pace' gait(frame 12) to the 'trot' gait(frame 345), and eventually to the 'gallop' gait(frame 678).}
    \label{snapshots}
\end{figure*}

Furthermore, our controller performs better in terms of energy efficiency and command tracking accuracy under different terrains and speed commands. In details, on relatively simple terrains with uniform noise, the energy efficiency of Complex Rewards increases rapidly with increasing speed. The energy efficiency of AMP remains low, but this is because it cannot track the high velocity commands well. And its energy efficiency is worse than our method in real world experiments, which demonstrates the advantage of our methods in terms of sim-to-real transfer. The learned controller can smoothly transit to different gait in order to respond to different velocity commands and different terrains, as shown in Fig.\ref{snapshots}.

\begin{table*}[htbp]
\captionof{table}{Comparison of Energy Efficiency and Command Tracking Accuracy}
\label{big_table}
\centering   
\begin{tabular}{|cc|ccc|cc|c|}
\hline
\multicolumn{2}{|c|}{terrain types} & \multicolumn{3}{c|}{uniform noise} & \multicolumn{2}{c|}{stairs} & \multicolumn{1}{c|}{discrete obstacles} \\ \hline
\multicolumn{2}{|c|}{controllers} & \multicolumn{1}{c|}{\begin{tabular}[c]{@{}c@{}}Complex \\ Rewards\end{tabular}} & \multicolumn{1}{c|}{AMP} & Ours & \multicolumn{1}{c|}{\begin{tabular}[c]{@{}c@{}}Complex\\  Rewards\end{tabular}} & Ours & Ours \\ 
\hline
\multicolumn{1}{|c|}{\multirow{2}{*}{0.5m/s} }& acc(\%) & \multicolumn{1}{c|}{$55.56\pm5.73$} & \multicolumn{1}{c|}{$46.37\pm3.21$} & $\bf{62.57\pm0.91}$ & \multicolumn{1}{c|}{$20.34\pm1.87$} & $\bf{54.99\pm1.21}$ & $\bf{57.84\pm1.06}$ \\ 
\cline{2-8} 
\multicolumn{1}{|c|}{} & pow(W) & \multicolumn{1}{c|}{$15.39\pm 9.36$} &   \multicolumn{1}{c|}{$\bf{13.75\pm7.32}$} &  $17.51\pm1.44$ &  \multicolumn{1}{c|}{$35.51\pm15.64$} &   $\bf{20.60\pm 5.94}$ &  $\bf{28.87\pm3.42}$ \\ 
\hline
\multicolumn{1}{|c|}{\multirow{2}{*}{1.0m/s}} & acc(\%) &  \multicolumn{1}{c|}{$53.49\pm7.83$} &  \multicolumn{1}{c|}{$45.39\pm2.09$} &  $\bf{62.55\pm0.65}$ &  \multicolumn{1}{c|}{$24.86\pm2.74$} &  $\bf{50.45\pm1.12}$ & $\bf{54.24\pm2.77}$ \\ 
\cline{2-8} 
\multicolumn{1}{|c|}{} & pow(W) &  \multicolumn{1}{c|}{$34.65\pm2.76$} & \multicolumn{1}{c|}{$\bf{19.48\pm0.80}$} &  $32.63\pm1.81$ & \multicolumn{1}{c|}{$50.18\pm35.44$} &  $\bf{36.11\pm1.57}$ & $\bf{50.93\pm19.00}$ \\ 
\hline
\multicolumn{1}{|c|}{\multirow{2}{*}{1.5m/s}} &  acc(\%) & \multicolumn{1}{c|}{$47.98\pm3.28$} & \multicolumn{1}{c|}{$39.08\pm4.67$} & $\bf{62.01\pm0.20}$ & \multicolumn{1}{c|}{\textbackslash{}} &  $\bf{30.28\pm2.75}$ & $\bf{40.64\pm3.67}$ \\ \cline{2-8} 
\multicolumn{1}{|c|}{} &
  pow(W) &
  \multicolumn{1}{c|}{$55.83\pm9.54$} &
  \multicolumn{1}{c|}{$\bf{41.99\pm10.49}$} &
  $42.42\pm1.44$ &
  \multicolumn{1}{c|}{\textbackslash{}} &
  $\bf{46.94\pm29.88}$ &
  $\bf{62.51\pm28.62}$ \\ \hline
\multicolumn{1}{|c|}{\multirow{2}{*}{2.0m/s}} &
  acc(\%) &
  \multicolumn{1}{c|}{$\bf{59.39\pm7.34}$} &
  \multicolumn{1}{c|}{$25.37\pm3.34$} &
  $54.89\pm1.47$ &
  \multicolumn{1}{c|}{\textbackslash{}} &
  $\bf{11.01\pm1.04}$ &
  $\bf{24.80\pm3.92}$ \\ \cline{2-8} 
\multicolumn{1}{|c|}{} &
  pow(W) &
  \multicolumn{1}{c|}{$79.52\pm23.04$} &
  \multicolumn{1}{c|}{$\bf{49.84\pm25.78}$} &
  $57.74\pm4.56$ &
 \multicolumn{1}{c|}{\textbackslash{}} &
  $\bf{65.80\pm36.96}$ &
  $\bf{69.43\pm37.20}$ \\ \hline
\multicolumn{1}{|c|}{\multirow{2}{*}{2.5m/s}} &
  acc(\%) &
  \multicolumn{1}{c|}{$44.50\pm 6.54$} &
  \multicolumn{1}{c|}{$9.34\pm 5.98$} &
  $\bf{53.85\pm1.48}$ &
  \multicolumn{1}{c|}{\textbackslash{}} &
  \multicolumn{1}{c|}{\textbackslash{}} &
  \textbackslash{} \\ 
  \cline{2-8} 
\multicolumn{1}{|c|}{} &
  pow(W) &
  \multicolumn{1}{c|}{$109.01\pm35.79$} &
  \multicolumn{1}{c|}{$\bf{9.33\pm5.93}$} &
  $67.24\pm9.60$ &
  \multicolumn{1}{c|}{\textbackslash{}} &
  \multicolumn{1}{c|}{\textbackslash{}} &
  \textbackslash{} \\ \hline
\end{tabular}
\vspace{-0.5cm}
\end{table*}

\subsection{Real World Experiments}
\textbf{Metrics}: 
In real world experiments, we compare the performance of our approach with the comparison baselines on following metrics: 
\begin{itemize}
\item\textbf{TTF}: Time to fall normalized by a threshold time. If robot does not fall during the threshold time, TTF is 1.
\item\textbf{Success rate}: The definition is same as Section~\ref{sim_exp}. To avoid damage to the robot, if the robot falls down in three consecutive experiments, its success rate is set to 0.
\item\textbf{Distance}: The distance the robot traverses in the threshold time, normalized by the desired distance. If the robot reach the desired distance in within the threshold time, then the distance is set to 1. 
\end{itemize}

\textbf{Results}: 
We evaluate the performance of controllers on the real robot in different settings as follows. Sample outcomes are shown in Figure~\ref{real_terrain}, videos can be found in the supplementary materials.
\begin{itemize}
    \item \textbf{Floating barrier high 13-cm}: We test the robot's capability of climbing over high obstacles placed on slippery ground. Both our controller and the baselines do not leverage any visual information. As for our controller, when the front legs are obstructed, one of the front legs immediately lifts upwards, while the other front leg extends, making the body to tilt upwards. Then the hind legs push with force to climb over the obstacle smoothly. This demonstrates the robot's capability to naturally switch gaits by utilizing past observations.    
    \item \textbf{Grassland}: There are many bumps and pits on the grassland. The robot with our controller can quickly adjust its posture to maintain balance when encountering obstacles, and move forward in walking and gallop gaits under different speed commands. It's worth mentioning that our method is the first to enabling quadrupedal robot to gallop in the wild.
    \item \textbf{Extremely slippery ground}: We create an extremely slippery surface by pouring oil onto plastic sheets. When the robot walks from the ground to the plastic sheet, it immediately adapts and switches to a more conservative gait, noticeably reducing its foot lift height, resulting in a gait resembling `glide' on the slippery surface. As a comparison, the robots with baseline controllers did not alter their gait on smooth surfaces, making it difficult to maintain balance and is easy to fall. 
    \item \textbf{Staircase high 8-cm}: Without visual information assistance, out method can successfully climb the stairs. This means our method enables the robot to continuously adjust its posture based on proprioceptive information. Furthermore, compared to previous method, our learned policy can climb the stairs with an agile and natural gait.
    \item \textbf{Slopes}: Although our robot has not been trained on slopes in simulation, it can successfully adapt to a slope of approximately 30 degrees. This demonstrates the generalization and robustness of our method.
\end{itemize}

\begin{table}[htbp]
\caption{Comparison of Success Rate and TTF}
\label{sucandttf}
\centering   
\begin{tabular}{|c|c|c|c|}
\hline
\makecell{terrain\\types}                       & controllers     & success rate & TTF  \\ \hline
\multirow{3}{*}{\makecell{unifor\\nois}}      & complex rewards & $0.973\pm0.004$          & $0.974\pm0.004$ \\ \cline{2-4} 
                                    & AMP             & $0.835\pm0.035$         & $0.848\pm0.039$ \\ \cline{2-4} 
                                    & ours            & $\bf{0.992\pm0.003}$         & $\bf{0.998\pm0.001}$    \\ \hline
\multirow{3}{*}{stairs}             & complex rewards & $0.521\pm0.101$         & $0.340\pm0.073$ \\ \cline{2-4} 
                                    & AMP             & $0\pm0$            & $0.248\pm0.031$ \\ \cline{2-4} 
                                    & ours            & $\bf{0.989\pm0.007}$         & $\bf{0.945\pm0.020}$ \\ \hline
\multirow{3}{*}{\makecell{discree\\obstacl}} & complex rewards & $0.607\pm0.113$         & $0.480\pm0.095$ \\ \cline{2-4} 
                                    & AMP             & $0.319\pm0.088$         & $0.363\pm0.060$  \\ \cline{2-4} 
                                    & ours            & $\bf{0.963\pm0.016}$         & $\bf{0.931\pm0.022}$ \\ \hline
\end{tabular}
\vspace{-0.4cm}
\end{table}

\begin{table}[H]
\caption{Results of Real World Experiments}
\label{real_result}
\centering   
\begin{tabular}{|c|c|c|c|c|}
\hline
terrain type                & controller                                                & success rate & TTF  & distance \\ \hline
\multirow{3}{*}{13-cm step} & \begin{tabular}[c]{@{}c@{}}Complex Reward\end{tabular} & 0.2          & 1    & 0.36     \\ \cline{2-5} 
                            & AMP                                                     & 0            & 1    & 0.2      \\ \cline{2-5} 
                            & Ours                                                      & \bf{0.8}          & \bf{0.88} & \bf{0.84}     \\ \hline
\multirow{3}{*}{grassland}  & \begin{tabular}[c]{@{}c@{}}Complex Reward\end{tabular} & 0.8          & 1    & 0.92     \\ \cline{2-5} 
                            & AMP                                                       & 0            & 0.1  & 0.1      \\ \cline{2-5} 
                            & Ours                                                      & \bf{1}            & \bf{1}    & \bf{1}        \\ \hline
\multirow{3}{*}{\begin{tabular}[c]{@{}c@{}}slippery\\ ground\end{tabular}} & \begin{tabular}[c]{@{}c@{}}Complex Reward\end{tabular} & 0 & 0.4 & 0.4 \\ \cline{2-5} 
                            & AMP                                                       & 0.2          & 0.6  & 0.68     \\ \cline{2-5} 
                            & Ours                                                      & \bf{0.8}          & \bf{0.9}  & \bf{0.94}     \\ \hline
\multirow{3}{*}{staircase}  & \begin{tabular}[c]{@{}c@{}}Complex Reward\end{tabular} & 0            & 0.98 & 0.56     \\ \cline{2-5} 
                            & AMP                                                       & 0            & 0.68 & 0.1      \\ \cline{2-5} 
                            & Ours                                                      & \bf{1}            & \bf{1}    & \bf{1}        \\ \hline
\end{tabular}
\end{table}

The quantitative results are shown in Table~\ref{real_result}, each data is an average over 10 experiments. We can see that our method outperforms all baselines in terms of the success rate, TTF and distance metrics. Note that the real world terrains are even more complex and diverse than that in simulation, with many unknown physical factors. Therefore, conducting real world experiments places high demands on the robustness of the controllers. Furthermore, we found that processing the output actions with a low-pass filter could smooth the motions to improve energy efficiency notably. We conducted ablation studies about the low-pass-filter for energy efficiency. As shown in Table~\ref{efficiency}, our approach exhibits better energy efficiency while tracking different velocity command, whether with the filter or not.

\begin{table}[H]
\caption{Comparison of Energy Efficiency}
\label{efficiency}
\centering   
\begin{tabular}{|cc|c|ccc|}
\hline
 &
   &
  \begin{tabular}[c]{@{}c@{}}Commanded Forward \\ Velocity (m/s)\end{tabular} &
  \begin{tabular}[c]{@{}c@{}}Complex\\ Reward\end{tabular} &
  AMP &
  Ours \\ \hline
\multicolumn{1}{|c|}{\multirow{4}{*}{\begin{tabular}[c]{@{}c@{}}Average \\ Power\\ (W)\end{tabular}}} &
  \multirow{2}{*}{\begin{tabular}[c]{@{}c@{}}w/\\ filter\end{tabular}} &
  0.5 &
  24.39 &
  15.62 &
  \bf{11.88} \\ \cline{3-6} 
\multicolumn{1}{|c|}{} &                                                                       & 1.0 & 41.03 & 33.85 & \bf{33.13} \\ \cline{2-6} 
\multicolumn{1}{|c|}{} & \multirow{2}{*}{\begin{tabular}[c]{@{}c@{}}w/o\\ filter\end{tabular}} & 0.5 & 25.88 & 20.66 & \bf{13.21} \\ \cline{3-6} 
\multicolumn{1}{|c|}{} &                                                                       & 1.0 & \bf{48.53} & 59.63 & 56.32 \\ \hline
\end{tabular}
\end{table}


\section{CONCLUSION}
In this paper, we proposed a novel framework which enables learning legged locomotion skills that are robust, agile, natural and energy efficient in real world challenging terrain environments. By adopting the teacher-student training paradigm, we obtained an encoder to infer environmental factors for adaptation in the real world. Several techniques are applied to enhance the robustness of the learned policy. Through the integration of an adversarial training branch, the learned gait is enforced to match the style of real animal motion capture data. Extensive results of a quadruped robot conducted in both simulation and real world showcased the superior performance of our proposed algorithm compared to baseline approaches. Future research could try to decouple the learning process for style matching and task completion.


\bibliographystyle{unsrt}
\bibliography{root}

\end{document}